\newcommand\cut[1]{}

\newcommand{\elbofinal}{\mathcal{L}}

\newcommand{\elbofinalfull}{\underline{\mathcal{L}}(\vtheta, \vgamma, \valpha)}
\newcommand{\squishlist}{
   \begin{list}{$\bullet$}
    { \setlength{\itemsep}{0pt}      \setlength{\parsep}{3pt}
      \setlength{\topsep}{3pt}       \setlength{\partopsep}{0pt}
      \setlength{\leftmargin}{1.5em} \setlength{\labelwidth}{1em}
      \setlength{\labelsep}{0.5em} } }

\newcommand{\squishlisttwo}{
   \begin{list}{$\bullet$}
    { \setlength{\itemsep}{0pt}    \setlength{\parsep}{0pt}
      \setlength{\topsep}{0pt}     \setlength{\partopsep}{0pt}
      \setlength{\leftmargin}{2em} \setlength{\labelwidth}{1.5em}
      \setlength{\labelsep}{0.5em} } }

\newcommand{\squishend}{
    \end{list}  }

{}
{}
{}
{}
{}
{}
{}

\newcommand{\rnd}[1]{\left(#1\right)}
\newcommand{\sqr}[1]{\left[#1\right]}

\newcommand{\myexpect}{\mathbb{E}}

\newcommand{\gauss}{\mbox{${\cal N}$}}

\newcommand{\myvec}[1]{\mbox{$\mathbf{#1}$}}
\newcommand{\myvecsym}[1]{\mbox{$\boldsymbol{#1}$}}

\newcommand{\valpha}{\mbox{$\myvecsym{\alpha}$}}

\newcommand{\veta}{\mbox{$\myvecsym{\eta}$}}
\newcommand{\vgamma}{\mbox{$\myvecsym{\gamma}$}}
\newcommand{\vmu}{\mbox{$\myvecsym{\mu}$}}

\newcommand{\vtheta}{\mbox{$\myvecsym{\theta}$}}

\newcommand{\vsigma}{\mbox{$\myvecsym{\sigma}$}}
\newcommand{\vSigma}{\mbox{$\myvecsym{\Sigma}$}}

\newcommand{\vg}{\mbox{$\myvec{g}$}}

\newcommand{\vm}{\mbox{$\myvec{m}$}}

\newcommand{\vx}{\mbox{$\myvec{x}$}}

\newcommand{\vz}{\mbox{$\myvec{z}$}}

\newcommand{\vF}{\mbox{$\myvec{F}$}}

\newcommand{\vH}{\mbox{$\myvec{H}$}}
\newcommand{\vI}{\mbox{$\myvec{I}$}}

\newcommand{\E}{\mbox{$E \;$}}

\newcommand{\be}{\begin{equation}}
\newcommand{\ee}{\end{equation}}
\newcommand{\bea}{\begin{eqnarray}}
\newcommand{\eea}{\end{eqnarray}}
\newcommand{\beaa}{\begin{eqnarray*}}
\newcommand{\eeaa}{\end{eqnarray*}}

\documentclass{article}

\usepackage[utf8]{inputenc}
\usepackage{algorithm} 
\usepackage{algpseudocode} 
\usepackage{mathtools}
\usepackage{xpatch}
\usepackage{enumitem}
\usepackage{hyperref}
\usepackage{movie15}
\usepackage{graphicx}
\usepackage{tikz}
\usepackage{amsmath, amsfonts}
\usepackage{dsfont}
\usepackage[square,sort,comma,numbers]{natbib}
\usepackage{mystyle}

\title{Natural Gradient Variational Inference with Gaussian Mixture Models
}

\begin{document}

\maketitle

\begin{abstract}
Bayesian methods estimate a measure of uncertainty by using the posterior distribution: $p(\vz|\mathcal{D}) = p(\mathcal{D}|\vz)p(\vz) / p(\mathcal{D})$.
One source of difficulty in these methods is the computation of the normalizing constant $p(\mathcal{D})= \int p(\mathcal{D}|\vz)p(\vz) dz$.
Calculating exact posterior is generally intractable and we usually approximate it. Variational Inference (VI) methods approximate the posterior  with a distribution $q(\vz)$ usually chosen from a simple family using optimization.\\ The main contribution of this work is described in section \ref{section3}, where we provide a set of update rules for natural gradient variational inference with mixture of Gaussians, which can be run independently for each of the mixture components, potentially in parallel. 
% Classical variational optimization is achieved through coordinate ascent which can be slow to converge.

% Variational bounds provide a convenient approach to approximate inference 
% Variational inference is one of the popular methods for approximating intractable posteriors $p(z|x)$ through optimization.
% Traditional variational inference methods involve 
% analytic computations and tedious bookkeeping for each new model. 
% The mean field algorithm requires analytical solutions of expectation with respect to approximate posterior, which are also intractable in general case.
\end{abstract}

\section {VI with Gradient Descent}
% Variational bounds provide a convenient approach to approximate inference. 
The recent approaches to VI generally avoid model-specific derivations. 
Black box variational inference \cite{BBVI} is one of the pioneer works that provides a recipe to quickly build and explore variety of models.

The idea is to assume a simple family for variational distribution $q_\phi(z|x)$ over the latent variables $z$ with parameters $\phi$, and find the member of that family that is closest in KL divergence to the true posterior. Note that if the generative model $p(z, x)$ is a parameterized model, we denote its parameters by $\theta$. Minimizing the KL divergence is an optimization problem, and its complexity is partially controlled by the complexity of the chosen variational family. 
\begin{flalign*}
     D_{KL}(q_\phi(z|x)||p_\theta(z|x)) &= E_{q_\phi(z|x)} [\log {q_\phi(z|x) \over p_\theta(z|x)}]\\
    %  &= E_{q_\phi(z|x)} [\log q_\phi(z|x) - \log p_\theta(z|x)]\\
     &= E_{q_\phi(z|x)} [\log q_\phi(z|x) - \log p_\theta(z,x) + \log p_\theta(x) ]\\
     &=    \log p_\theta(x) + E_{q_\phi(z|x)}[ \log q_\phi(z|x) - \log p_\theta(z,x) ]\\
\end{flalign*}

The term $E_{q_\phi(z|x)}[ \log {p_\theta(z,x) -\log q_\phi(z|x)}] $ is called the variational lower bound or alternatively, the Evidence Lower BOund (ELBO). We denote the ELBO by $\mathcal{L}$.
% We denote the ELBO by $\mathcal{L}(\theta, \phi; x)$, and re-write it in a few different ways as follows:
% \begin{align}
%      \mathcal{L}(\theta, \phi; x) &= E_{q_\phi(z|x)}[\log p_\theta(z,x)] - E_{q_\phi(z|x)}[\log q_\phi(z|x)]\\ \label{elbo1}
%      &=  E_{q_\phi(z|x)}[\log p_\theta(x|z)]+E_{q_\phi(z|x)}[\log p_\theta(z)] -E_{q_\phi(z|x)}[ \log q_\phi(z|x)]\\
%      &= E_{q_\phi(z|x)}[\log p_\theta(x|z)] -D_{KL} (q_\phi(z|x) || p_\theta(z)) \label{elbo2}
% \end{align}
% Intuitively, the first term of equations \ref{elbo1} and \ref{elbo2}  favours distributions $p(x,z)$ that place high probability mass on $z$'s that also explain the observations. The second term in equation \ref{elbo1}  rewards variational distributions that are entropic and maximize uncertainty by spreading their mass on many different configurations of $z$, while the second term of \ref{elbo2}
% is the negative divergence between the variational density and the prior; it encourages densities close to the prior.
Since the log-likelihood of data, $p(x)$ is assumed to be fixed, minimizing the KL divergence is equivalent to maximizing the ELBO.
\begin{align}
    \log p_\theta(x)= D_{KL}(q_\phi(z|x)||p_\theta(z|x)) & + E_{q_\phi(z|x)}[ \log {p_\theta(z,x) -\log q(z|x)}]
\end{align}
In \cite{BBVI}, the gradient of ELBO is written as the expectation with respect to $q_\phi$. Therefore, we can approximate it using Monte Carlo samples from $q_\phi$. This gives us an unbiased estimator for the gradient of ELBO. 
\begin{align}
    \nabla_{\phi}\mathcal{L} &=\nabla_{\phi} E_q {(\log p(x,z)-\log q(z|\phi))}\\
    &= E_q[(\log p(x,z)-\log q(z|\phi))\nabla_{\phi} \log q(z|\phi)]
\end{align}
A noisy unbiased gradient estimator of ELBO with Monte Carlo samples $z^{(l)}$ from variational distribution is obtained as follows:
\begin{align}
    \nabla_{\phi}\mathcal{L} \approx 
    {1 \over L} \sum_{l=1}^L (\log p_\theta(x,z^{(l)})-\log q(z^{(l)}|\phi)) \nabla_{\phi} \log  q(z^{(l)}|\phi)
\end{align}
This gradient estimate can be used by a stochastic optimizer to achieve the optimized values for variational parameters $\phi$. In stochastic optimization a function is optimized using noisy estimates of its gradient.
Here Monte Carlo estimates are the noisy estimates of the gradient.
This estimator is unbiased but it exhibits high variance. 
\section{VI with Natural Gradient Descent}
Natural gradient descent assumes the parameter of interest lies on a Rimannian manifold and selects the steepest direction along that manifold.
These methods exploit the Riemannian geometry of the approximate posterior and scale the gradient by the inverse of metric tensor. In case of parametric families, the Fischer Information matrix (FIM) induces a Rimannian manifold. 
We usually intend to avoid direct computation of FIM. \\
Suppose we aim to optimize a variational objective $\mathcal{L}(\eta)$ as follows:
\begin{align}
\mathcal{L}(\veta) &= \mathds{E}_{q_\eta(z)} \sqr{ \log p(z) - \log q(z) +  \sum_{i=1}^N \sqr{\log(p(D_i|z)} } 
\end{align}

{\bf \noindent Exponential Family.} Assume the variational distribution  $q_\eta(\vz)$ has a minimal exponential family form with natural parameters $\veta$. 
Then, there exist a one-to-one mapping between the natural parameters $\veta$ and expectation parameter $\vm$. 
Therefore, we can rewrite the variational objective $\mathcal{L}(\veta)
$ as a function of expectation parameters $\vm$ instead of natural parameters $\veta$. 
\begin{align}
   \mathcal{L_*}(\vm) \coloneqq \mathcal{L}(\veta) 
\end{align}

{\bf \noindent Theorem 1. Variable Transformation.} In exponential family distributions, the natural gradient wrt to natural parameters $\veta$ is equal to gradient wrt expectation parameters $m$.
\begin{align}
    F(\veta)^{-1} \nabla_{\veta} \mathcal{L}(\veta) = \nabla_{\vm} \mathcal{L}_* (\vm)
\end{align}
{\bf Proof:} The proof is based on \cite{mirror}. We can write the derivative wrt $\veta$  in terms of $\vm$ using the chain rule:
\begin{align}
     \nabla_{\veta} \mathcal{L}(\veta) = \sqr{\nabla_{\eta} \vm} \nabla_{\vm} \mathcal{L}_* (\vm) = F(\veta) \nabla_{\vm} \mathcal{L}_* (\vm) 
\end{align}
The second equality comes from the fact that in exponential family distributions, natural parameters $\veta$ and expectation parameters $\vm$ are related through 
Legendre transform: $\vm = \nabla A(\eta)$ where $A(\veta)$ is the log-partition function. Therefore, $\nabla_{\eta} \vm = \nabla^2_{\eta \eta} A(\veta)$. By the definitions of the FIM, we have 
$F(\veta) \coloneqq  \nabla^2_{\eta \eta} A(\veta)$. So $\nabla_{\eta} \vm =F(\veta)$. \\
It is worth mentioning that FIM in one space is the inverse of FIM in the other space. This enables computing natural gradient in one space using the gradient in the other. \\

{\bf \noindent Theorem 2. Mirror Descent.} Mirror descent induced by a Bregman
divergence proximity functions is equivalent to the natural gradient descent algorithm on the dual Riemannian manifold.\\
{\bf Proof:} 
Assume we have a strictly convex $G$ that induces the Bregman Divergence $B_G:\Theta \times \Theta \rightarrow {R}^+$, where $B_G$ is defined as follows:
\begin{align}
    B_G(m, m_t) = G(\vm) - G(\vm') - \langle \nabla G(\vm'), \vm - \vm' \rangle
\end{align}
The mirror descent update is written as follows:
\begin{align}
\vm_{t+1} = \text{argmin}_{\vm} \langle \vm, \nabla \mathcal{L_*}(\vm_t) \rangle + {1 \over \alpha_t} B_G(\vm, \vm_t)
\end{align}
Taking the derivative to find the minimum, we arrive at:
\begin{align}
    \nabla_{\vm} G({\vm}_{t+1}) = \nabla_{\vm} G({\vm}_t) - \alpha_t \nabla_{\vm} \mathcal{L_*}(\vm_t)
\end{align}
In terms of the dual variable $\veta=\nabla G(\vm)$ and noting that $\vm = \nabla H(\veta)$,
\begin{align}
    \veta_{t+1} = \veta_t - \alpha_t \nabla_{\vm} \mathcal{L_*}(\nabla H(\veta_t))
\end{align}
Applying chain rule, we have:
\begin{align}
    \nabla_{\vm} \mathcal{L_*}(\nabla H(\veta_t)) = \sqr{\nabla_{\veta} \nabla H(\veta_t)}^{-1}  \nabla_{\veta} \mathcal{L_*}(\nabla H(\veta_t))
\end{align}
Therefore:
\begin{align}
    \veta_{t+1} = \veta_t - \alpha_t \sqr{\nabla^2 H(\veta_t)}^{-1} \nabla_\eta \mathcal{L_*}(\nabla H(\veta_t))
\end{align}
In our settings, the function $G$ is the negative entropy function: $ G(m) = \sum{q_m(\theta) \log q_m(\theta)}$. The mirror descent update can be written as follows: 

\begin{align}
 \vm_{t+1} = \text{argmin}_{\vm} \langle \vm, \nabla_{m} \mathcal{L}_*(\vm_t) \rangle + {1 \over \beta} \mathcal{D}_{KL} \sqr{q_m(\theta) || q_{m_t} (\theta)}
\end{align} 

In this case the function $H$ which is the convex conjugate of $G$, is the  
Mirror descent and natural gradient descent are both generalizations of online gradient descent when the parameter of interest lies on a non-Euclidean manifold.
% the natural-gradient update can be performed using the gradient with respect to the expectation parameter.
In this section we showed that mirror descent update in the expectation parameters space is equivalent to the natural gradient update in the natural parameter space. \\

Each step of this mirror descent update is equivalent to the following natural gradient descent in the natural parameter space. 
\begin{align}
    \veta_{t+1} = \veta_{t} + \beta_{t} \vF(\veta_{t})^{-1} \nabla_{\eta}\mathcal{L}(\eta_{t}) \label{eq1}
\end{align}
Using Theorem 1, we can now write (\ref{eq1}) as follows:
\begin{align}
    \veta_{t+1} = \veta_{t} + \beta_{t} \nabla_{m} \mathcal{L}_* (\vm_{t}) \label{eq2}
\end{align}
\subsection{NGVI with Gaussian Mean-Field Approximation}
Natural gradient descent assumes the parameter of interest lies on a Riemannian manifold and selects the steepest descent direction along that manifold \cite{amarinatural}. 
This section contains natural gradient updates for Gaussian approximate posterior $q_\eta(\vtheta) \coloneqq \mathcal{N}(\vtheta|\vmu, \vSigma)$ with mean $\vmu$ and covariance matrix $\vSigma$. 
Natural and expectation parameters of a Gaussian are defined as follows:
\begin{align}
    \veta^{(1)} \coloneqq \Sigma^{-1}\vmu ,\; &\veta^{(2)} = -{1\over 2} \vSigma^{-1}  \label{eta_gaussian}\\
     \vm^{(1)} \coloneqq E_{q}[\vtheta] = \vmu , \; &\vm^{(2)} = E[\vtheta \vtheta ^T] = \vmu \vmu^T + \vSigma
\end{align}
Gradient of $\mathcal{L}_*$ wrt expectation parameters:
\begin{align}
    % &q(z_n) = \mathcal{N}(z_n|\mu, \Sigma) \\
    % &m^{(1)} = \mu , M^{(2)} = \Sigma + \mu^2 \\
    &\nabla_{m^{(1)}} \mathcal{L} =  \nabla_{\mu}{\mathcal{L}} . {\nabla_{m^{(1)}}{\mu}} + {\nabla_{\Sigma} \mathcal{L} . {\nabla_{m^{(1)}} \Sigma}} = {{\nabla_\mu \mathcal{L}}} -2 {\nabla_\Sigma \mathcal{L}} .\mu \\
     &\nabla_{m^{(2)}}.\mathcal{L} =  \nabla_{\mu}{\mathcal{L}} .{\nabla_{m^{(2)}}\mu  }
     + {\nabla_{\Sigma} \mathcal{L}}. {\nabla_{m^{(2)}} \Sigma} = {\nabla_\Sigma \mathcal{L}}
    % { \partial{L} \over \partial {d}}
\end{align}
We can rewrite $\nabla_{m} \mathcal{L}_*$ in  (\ref{eq2}) in terms of $\vmu$ and $\vSigma$ by substituting  natural parameters $\veta^{(1)}$ and $\veta^{(2)}$ with their corresponding values 
for Gaussian distribution as shown in (\ref{eta_gaussian}). 
\begin{align}
    &\vSigma_{t+1}^{-1} = \vSigma_{t}^{-1}-2\beta_t[\nabla_\Sigma \mathcal{L}] \\
    &\vm_{t+1} = \vmu_t + \beta_t \Sigma_{t+1} [\nabla_\mu \mathcal{L}]
\end{align}

\subsection{Variational Online Newton(VON)}
We can reformulate the NGVI updates in terms of gradient and Hessian of negative log-likelihood of data, defined as $f(z) = \sum_{i=1}^N \log(p(D_i|z)$ as follows \cite{vadam}:
\begin{align}
    \nabla_\mu \elbofinal &= \nabla_\mu \myexpect_q \sqr{\log p(z) - \log q(z) + f(z)}\\
    &= \lambda \mu + \myexpect_q \sqr{\vg(z)}\\
    \nabla_\Sigma \elbofinal &= \nabla_\Sigma \myexpect_q \sqr{\log p(z) - \log q(z) + f(z)}\\
    &= -{1 \over 2}\lambda\vI +{1 \over 2} \Sigma^{-1}+{1 \over 2} \myexpect_q \sqr{\vH(z)},
\end{align}
where $\vg(z)$ and $\vH(z)$ denote the gradient and Hessian of negative log-likelihood respectively, and are obtained using Bonnet's and Price's theorem \cite{opper2009variational} as shown below:
\begin{align}
    \nabla_\mu \myexpect_q \sqr{f(z)} &= \myexpect_q \sqr{\nabla_z f(z)} \coloneqq \myexpect_q \sqr{\vg(z)}\\
    \nabla_\Sigma \myexpect_q \sqr{f(z)} &= \myexpect_q \sqr{\nabla^2_{zz} f(z)} \coloneqq {1 \over 2}\myexpect_q \sqr{\vH(z)}
\end{align}
We estimate each of the expectations above with one Monte-Carlo sample $z_0 \sim \mathcal{N}(z| \mu_t, \Sigma_t)$, and arrive at the update rules of VON \cite{vadam}:
\begin{align}
    \mu_{t+1} &= \mu_t - \beta \Sigma_{t+1} \sqr{\vg(z_0) + \lambda \mu_t}\\
    \Sigma^{-1}_{t+1} &= (1-\beta) \Sigma^{-1}_t + \beta \sqr{\vH(z_0)+\lambda \vI}
\end{align}

\subsection{Variational Online Gauss-Newton(VOGN)}
VOGN is similar to VON, however the Generalized Gauss-Newton approximation is used to estimate the Hessian:
\begin{align}
    \nabla^2_{z_j z_j} f(\vz) \approx{{1 \over |M|} \sum_{i \in M} \sqr{\nabla_{z_j} f_i(\vz)}^2}
\end{align}
where $z_j$ is the j'th element of $\vz$, and $M$ is the minibatch with size $|M|$.

\section{NGVI with Mixture of Gaussians}
Gaussian mixture models are powerful approximation where components are mixed using a discrete distribution. 
This distribution can not be written in an exponential form in general. In this section, we demonstrate the update rules for a mixture of Gaussians based on \cite{lin2019fast}. 
We also include two solutions to convert the algorithm in a way that sampling from MoG can be replaced by sampling from each of the Gaussian components independently.
As a result, this part can be run in parallel for speed up.

\begin{algorithm}
	\caption{NGVI for MoG (\cite{lin2019fast})} 
	\begin{algorithmic}[1]
	\State Mixture of Gaussians with $K$ components. \\ For each component $c$, mean: $\mu_c$, variance: $\Sigma_c$, mixing proportion: $\pi_c$
	\State $ h(z) \coloneqq \bigg[ \log {q(z) - \log p(z)}  - \sum_{n=1}^N [\log(p(D_n|z)] \bigg]$
	\State {ELBO objective: $\mathcal{L} = \mathds{E}_{q(z)}[-h(z)]$} 
	\For {$e \leftarrow 1: \text{MaxEpochs }$}
		
		  %  \State $ z_0 \sim  q(z, w)$
		    \State $i \sim Categorical \{\pi_1,\pi_2,\ldots,\pi_K\}$ \Comment{Sample a component wrt. $\pi$'s}
		    \State $\vm{z_0} \sim \mathcal{N}(\mu_i, \Sigma_i)$  \Comment{Sample from i'th component}
		    \For {component $c$ in $\{1,2,\ldots, K\}$}
			\State $\delta_c(z_0) \coloneqq \mathcal{N}(z_0|\mu_c, \Sigma_c)/ \sum_{c'=1}^K {\pi_{c'} \mathcal{N}(z_0|\mu_{c'}, \Sigma_{c'})}$
            \State $[\Sigma_c^{(new)}]^{-1} \leftarrow \Sigma_c^{-1}  + \beta  \delta_c(z_0) [\nabla_z^2 h(z_0)]$
            \State $[\mu_c^{(new)}] \leftarrow \mu_c - \beta   \Sigma_c^{(new)} \delta_c(z_0)[\nabla_z h(\vm{z_0})]$
		\EndFor
		\For {component $c$ in $\{1,2,\ldots, K\}$}
		    \State $\rho_c \coloneqq \log (\pi_c / \pi_K)$
		    \State $\rho_c \leftarrow \rho_c - \beta(\delta_c(z_0) - \delta_K(z_0)) h(\vm{z_0})$
		\EndFor
		\State $\vm{\rho} = \vm{\rho} - \max(\vm{\rho})$
        \State $\vm{\pi} = softmax(\vm{\rho})$
		
	\EndFor
	
	\end{algorithmic} 
\end{algorithm}

\subsection{Alternative set of update rules of NGVI for Mixture of Gaussians}\label{section3}
Let's rewrite the ELBO as follows:
\begin{align}
\mathcal{L} &= \mathds{E}_{q(z)}\left[\log p(z) - \log q(z) +  \sum_{n=1}^N \left[\log(p(D_n|z)\right]  \right]  \\
% &= \mathds{E}_{q(z)} \sqr{\log p(z) +  \sum_{n=1}^N \log(p(D_n|z)  }  - \mathds{E}_{q(z)}\sqr{ \log q(z)} \\ 
% &= \mathds{E}_{q(z)}\left[\log p(z) +  N f(z) \right]  - \mathds{E}_{q(z)}\left[ \log q(z)\right] \\ 
&= \mathds{E}_{q(z)}\left[t(z)\right] - \mathds{E}_q(z)\left[\log q(z)\right]
\end{align}
Where $t(z) = \log p(z) +  N f(z)] $ and $f(z) = -{1 \over N}\sum_{n=1}^N [\log(p(D_n|z) $. 
Here the approximate posterior distribution is MoG: $q(z) = \sum_{c=1}^K {\pi_{c} \mathcal{N}(z|\mu_{c}, \Sigma_{c}) }$.
Now we take the derivative of $\mathcal{L}$ with respect to MoG parameters analytically. 
\subsubsection{Derivation of the gradient of $\mathcal{L}$ wrt $\mu_c$}
\begin{align}
\nabla_{\mu_c} \mathcal{L} &= \nabla_{\mu_c} \mathds{E}_{q(z)}\left[t(z) - \log q(z)\right] \\
&=\mathds{E}_{q(z)} \sqr{q(w=c|z) \nabla_z \rnd{ t(z) - \log q(z)}}\\
&=\pi_c \mathds{E}_{q(z)}\sqr{\delta_c(z) \nabla_z \rnd{t(z)- \log q(z)}}
\end{align}
The last line results from Bonnet and Price theorem \cite{opper2009variational} for GMMs. 
In order to calculate $\nabla_z \log q(z)$, we use the log-derivative trick:
\begin{align}
 \nabla_z \log q\left(z\right) &= {1 \over q(z)} \nabla_z q(z) \label{log_dev}\\
 &=-\sum_c \pi_c \delta_c(z)* {\left({{z-\mu_c} \over \sigma_c ^2} \right)} 
\end{align}
Please note that  $\delta_c(z) \coloneqq \mathcal{N}(z|\mu_c, \Sigma_c)/ \sum_{c'=1}^K {\pi_{c'} \mathcal{N}(z|\mu_{c'}, \Sigma_{c'})}$
Also note that  $\nabla_z q(z)$ can be obtained as follows:

\begin{align}
 \nabla_z q(z) &= \nabla_z  \sum_c \pi_c \mathcal{N}\left(\vz| \mu_c, \sigma_c \right) \\
 &= -\sum_c \pi_c \mathcal{N}\left(\vz| \mu_c, \sigma_c\right) * {\left({{z-\mu_c} \over \sigma_c ^2} \right)}  \label{last_eq_dev} 
\end{align} 
% We sample $z_0 \sim q(z)$
As defined above, $t(z) = \log p(z) +  N f(z) $.\\
 $p(z)$ is chosen to be Gaussian with diagonal covariance: $p(z) = \mathcal{N}(z|0, {I / \lambda})$. We define  $g(z)  \coloneqq \nabla_z f(z)$.
\begin{align}
\nabla_{\mu_c} \mathcal{L} &=\pi_c \mathds{E}_{q(z)} \sqr{\delta_c(z) \left( \nabla_z t(z) + \sum_j \pi_j \delta_j(z)* {({{z-\mu_j} \over \sigma_j ^2} )}\right) }\\
&=\pi_c \int N(z|\mu_c, \sigma_c) \left[Ng(z) - \lambda \mu_c + \sum_j \pi_j \delta_j(z)* {({{z-\mu_j} \over \sigma_j ^2} )} \right] dz \\
&=\pi_c \mathds{E}_{ N(z|\mu_c, \sigma_c)} \left[Ng(z) - \lambda \mu_c + \sum_j \pi_j \delta_j(z)* {({{z-\mu_j} \over \sigma_j ^2} )} \right] \\
\end{align}
Using this solution, we could rewrite the derivation with Gaussian sampling instead of MoG sampling. 
The update rule for $\mu_c$ can be written as follows:
 \begin{align}
 \mu_c^{(new)} &\leftarrow \mu_c + \beta {1 \over \pi_c} \left(\nabla_{\mu_c} \mathcal{L}\right)\\
 \mu_c^{(new)} &\leftarrow \mu_c + \beta\mathds{E}_{ N(z|\mu_c, \sigma_c)} \left[Ng(z) - \lambda \mu_c + \sum_j \pi_j \delta_j(z)* {\left({{z-\mu_j} \over \sigma_j ^2} \right)} \right]  \\
  \mu_c^{(new)} &\leftarrow \left(1-\beta \lambda \right)\mu_c + \beta \left(  Ng(z_0)  + \sum_j \pi_j \delta_j(z_0)* { \left( {{z_0-\mu_j} \over \sigma_j ^2} \right)} \right)
 \end{align}
In the last line, $z_0$ is a sample from component $c$ of MoG: $z_0 \sim \mathcal{N}(z|\mu_c, \sigma_c)$
 \subsubsection{Gradient of $\mathcal{L}$  wrt $\sigma_c$}
 \begin{align}
\nabla_{\sigma_c} \mathcal{L} &= \nabla_{\sigma_c} \mathds{E}_{q(z)}\sqr{t(z) - \log q(z)} \\
&=\mathds{E}_{q(z)} \sqr{q(w=c|z) \nabla_z^2 \rnd{ t(z) - \log q(z)}}\\
&=\pi_c \mathds{E}_{q(z)}\sqr{\delta_c(z) \nabla_z^2 \rnd{t(z)- \log q(z)}}
\end{align}

 Using log-derivative trick, we have: 
 \begin{align}
 \nabla_z^2 \log q(z) &= \nabla_z [{1 \over q(z)} \nabla_z q(z)]\\
%  &=-({\nabla_z q(z) \over q(z)})^2  + {1 \over q(z)} * \nabla_z^2 q(z)\\
 &= (\sum_c \pi_c \delta_c(z) * {-({{z-\mu_c} \over \sigma_c ^2} )} )^2 +  {1 \over q(z)} * \nabla_z^2 q(z)\\
 \end{align}We need to derive $\nabla_z^2 q(z)$:
\begin{align}
\nabla_z^2 q(z) &= 
% \nabla_z (\nabla_z q(z)) \\
% &= \nabla_z (\sum_c \pi_c \mathcal{N}(\vz| \mu_c, \sigma_c) * {-({{z-\mu_c} \over \sigma_c ^2} )} )\\
% &= \sum_c \pi_c \nabla_z  (\mathcal{N}(\vz| \mu_c, \sigma_c) * {-({{z-\mu_c} \over \sigma_c ^2} )} )\\
\sum_c \pi_c ( \mathcal{N}(\vz| \mu_c, \sigma_c) * {({{z-\mu_c} \over \sigma_c ^2} )}^2 -  \mathcal{N}(\vz| \mu_c, \sigma_c) *{1 \over \sigma_c^2})\\
&= \sum_c \pi_c [ \mathcal{N}(\vz| \mu_c, \sigma_c) * ({({{z-\mu_c} \over \sigma_c ^2} )}^2 - {1 \over \sigma_c^2})]\\
\end{align}
Then we have:
 \begin{align}
 \nabla_z^2 \log q(z)   &=-({\nabla_z q(z) \over q(z)})^2  + {1 \over q(z)} * \nabla_z^2 q(z)\\
%  &= (\sum_c \pi_c \delta_c(z) * {-({{z-\mu_c} \over \sigma_c ^2} )} )^2 + \sum_c \pi_c \delta_c(z) * \left({({{z-\mu_c} \over \sigma_c ^2} )}^2 - {1 \over \sigma_c^2} ) \right)\\
 &= ( \nabla_z \log q(z) )^2 +  \nabla_z \log q(z) * {({{z-\mu_c} \over \sigma_c ^2} )}-  \sum_c  { \pi_c \delta_c(z) \over \sigma_c^2}
 \end{align}
Therefore:
\begin{align}
\nabla_{\sigma_c} \mathcal{L} &=\pi_c \mathds{E}_{q(z)}\sqr{\delta_c(z)\nabla_z^2 \rnd{t(z) - \log q(z)}}\\
&=\pi_c \mathds{E}_{\mathcal{N}(z|\mu_c, \sigma_c)} [\nabla_z^2 \rnd{t(z)-\log q(z)}]
\end{align}
The update rule for $\sigma_c$ can be written as follows:
\begin{align}
-{1 \over 2} [\sigma_c^{(new)}]^{-1} &\leftarrow -{1 \over 2} [\sigma_c]^{-1} + {\beta \over \pi_c}(\nabla_{\sigma_c} \mathcal{L})\\
-{1 \over 2} [\sigma_c^{(new)}]^{-1} &\leftarrow -{1 \over 2} [\sigma_c]^{-1} + \beta \mathds{E}_{\mathcal{N}(z|\mu_c, \sigma_c)} \sqr{\nabla_z^2 [t(z)-  \log q(z)]}
\end{align}

\subsubsection{Gradient of $\mathcal{L}$  wrt $\pi_c$}
The gradient of the variational distribution wrt $\pi_c$ is as follows:

\begin{align}
    \nabla_{\pi_c} q(\vz) = \nabla_{\pi_c} \sum_{k=1}^K \pi_kq(\vz|w=k) = q(\vz|w=c) - q(\vz|w=K),
\end{align}
where $q(\vz|w=c)=\mathcal{N}(\vz|\mu_c,\sigma_c)$. Please note that the second term is resulted because we have $\pi_K = 1- \sum_{c=1}^{K-1} \pi_c$. Also recall that $t(z) = \log p(z) +  N f(z)] $ where $f(z) = -{1 \over N}\sum_{n=1}^N [\log(p(D_n|z) $.

\begin{align}
\nabla_{\pi_c} \elbofinal &= \nabla_{\pi_c} \myexpect_{q(z)}{\sqr{t(z)- \log q(z)}}\\
&= \int{\nabla_{\pi_c} q(z) \sqr{t(z)-\log q(z)}} d\vz - \int{q(\vz) \nabla_{\pi_c} \log q(\vz) d\vz}\\
&= \int{\sqr{q(\vz|w=c) - q(\vz|w=K)} \sqr{t(z)-\log q(\vz)} d\vz}\\
&= \myexpect_{\mathcal{N}(z|\mu_c, \sigma_c)} \sqr{-h(z)} +  \myexpect_{\mathcal{N}(z|\mu_K, \sigma_K)} \sqr{-h(z)}\\
&= -\sqr{q(z_c|w=c)h(z_c)} - \sqr{q(z_K|w=K) h(z_K)}
\end{align}
The update can be performed as follows:
\begin{align}
    \log ({\pi_c \over \pi_K}) \leftarrow  \log ({\pi_c \over \pi_K}) + \beta\sqr{q(z_c|w=c)h(z_c) + q(z_K|w=K) h(z_K)}
\end{align}

In the last equation, we need to draw two [sets of] samples from two distributions: $z_c \sim \mathcal{N}(\mu_c, \sigma_c)$ and $z_K \sim \mathcal{N}(\mu_K, \sigma_K)$.

\section{The Entropy Trick}
The following entropy trick will reduce the variance of our estimator:
\begin{align*}
     H[\vz]=-\sum_c \pi_c \log \sum_j \pi_j q(\hat{\vz}|w=j)+H[\vz|w] - \hat{H}[\vz|w],
\end{align*}
where $H[\vz|w]$ is calculated using analytical definition of Gaussian entropy, and $\hat{H}[\vz|w]$ is sample estimator: \\
\begin{flalign}
 H[\vz|w] &=  \sum \pi_c H[\vz|w=c] =  \sum \pi_c ({1\over 2} \log 2\pi e\sigma_c^2 ).\\
\hat{H}[\vz|w] &= \sum \pi_c H[\vz|w=c] \\
&= \sum_c \pi_c \E_{q(z|w=c)}[-\log q(\vz|w=c)] \\
&=\sum_c \pi_c/S (-\sum^S_{s=1}\log q(\vz_s|w=c)) \\
& \rnd{z_s \sim q(\vz|w=c)}
\end{flalign}
 
\subsection{Proof of the Entropy Trick}
We define a variational distribution $q(z)$ as a mixture distribution as follows:
\begin{align}
    q(\vz) = \sum_c \pi_c q(\vz|w=c) \\
    q(\vz|w=c) = \gauss(\vmu_c, \vsigma_c)
\end{align}

Using Bayes for entropy, the entropy term can be written as:
\begin{align*}
%     H[\vz]+H[w|\vz]&= H[w]+H[\vz|w]  \\
    H[\vz] = H[w]+H[\vz|w]-H[w|\vz]
\end{align*}
 
Each term of the right hand side of equation mentioned above can be computed as shown below.\\
1) $H[w]$: This entropy term is given by definition as follows:

\begin{align}
    H[w] &= -\sum_c \pi_c \log \pi_c
\end{align}
2) $H[\vz|w]$:
\begin{align}
    H[\vz|w] &= \myexpect{E}_{q(z,w)}[- \log q(z|w)] \\
    % &= \E_{q(w)}[\E_{q(z|w=c)} [-\log q(z|w=c)]]  \\
    % &= \sum_c \pi_c \E_{q(z|w=c)}[-\log q(z|w=c)]\\
    % &=\sum_c \pi_c H[z|w=c]
\end{align}\\
The last line results from the analytical definition of entropy being the weighted sum of component's entropy: $\sum \pi_c H[\vz|w=c]$. \\
3) $H[w|\vz]$:
\begin{align}
    H[w|\vz] &= H[w] + H[\vz|w] - H[\vz] \\
%     &= \myexpect_{q(\vz,w)}[- \log q(w|\vz)] \\
%     &=  \E_{q(z,w)}[-\log q(w) - \log q(\vz|w) + \log q(\vz)]  \label{second_eq_1}\\ 
    &=  H[w] +\sum_c \pi_c \E_{q(z|w=c)}[-\log q(\vz|w=c)] \\
    &+\sum_c \pi_c E_{q(z|w=c)}[\log \sum_j \pi_j q(\vz|w=j)]  
%      + \sum_c \pi_c \E_{q(\vz|w=c)}[-\log q(\vz|w=c)+\log \sum_j \pi_j q(\vz|w=j)]  \label{last_eq_1}\\ 
%      &=H[w]+\sum_c \pi_c \E_{q(\vz|w=c)}[-\log q(\vz|w=c)] +E_{q(\vz|w=c)}[\log \sum_j \pi_j q(\vz|w=j)]  
\end{align}
The last two terms will be evaluated by sampling from each component. 
Denoting sample estimators $\hat{z}$ and $\hat{H}$, we have:
\begin{align}
    H[w|\vz] = H[w] + \hat{H}[\vz|w] + \sum_c \pi_c \log \sum_j \pi_j q(\hat{\vz}|w=j) \label{trick_eq}
\end{align}
Therefore by plugging (\ref{trick_eq}) we have:
\begin{align}
    H[\vz] &= H[\vz|w] + H[w] - H[w|\vz] \\
    &= H[\vz|w] + H[w] - H[w] - \hat{H}[\vz|w] - \sum_c \pi_c \log \sum_j \pi_j q(\hat{\vz}|w=j)\\
    &=H[\vz|w] - \hat{H}[\vz|w] -\sum_c \pi_c \log \sum_j \pi_j q(\hat{\vz}|w=j)
\end{align}
Note $\E_{q(\vz|w=c)}[-\log q(z|w=c)]$ could be computed analytically, but we sample it because 
it will act as a control variate to reduce the variance of the gradient.

The marginal likelihood can be expanded as: $p(\vx) = \int p(\vx|\vz)p(\vz) d\vz$, where  $p(\vz) = \mathcal{N}(0,I)$. 
Also the variational distribution is assumed to be a MoG: $q(\vz) = \sum_c \pi_c q(\vz|w=c)$. 
Noting that $\elbofinal$ is the ELBO, then we have:

\begin{align}
    \log p(\vx) &= \log \int p(x|\vz)p(z)dz \\
     &= \int q(\vz) \log p(\vx|\vz) d\vz + \int q(\vz) \log p(\vz) - \int q(\vz) \log q(\vz)\\
    &=\E_{q(z)}[\log p(\vx|\vz)] + \E_{q(z)}[\log p(\vz)] + H[\vz] = \elbofinal
\end{align}
Using the trick, we can write the $\elbofinal$ as follows:
\begin{align*}
\elbofinal
&=\E_{q(\vz)}[\log p(x|\vz)] + \E_{q(\vz)}[\log p(\vz)]\\
&+H[\vz|w] - \hat{H}[\vz|w] - \sum_c \pi_c \log \sum_j \pi_j q(\hat{\vz}|w=j)\\
&=\sum_c \pi_c [E_{q(\vz|w=c)}[\log p(x|\vz)] + E_{q(\vz|w=c)}[\log p(\vz)-\log q(\vz|w=c)]\\
&- H[q(\hat{\vz}|w=c)] - \log \sum_j \pi_j q(\hat{\vz}|w=j)\\
&=\sum_c \pi_c [E_{q(\vz|w=c)}[\log p(x|\vz)] - KL[q(\vz|w=c)  ||  p(\vz)]\\
&- H[q(\hat{\vz}|w=c)] - \log \sum_j \pi_j q(\hat{\vz}|w=j)\\
&=\sum_c \pi_c [\elbofinalfull_c - H[q(\hat{\vz}|w=c)]-\log \sum_j \pi_j q(\hat{\vz}|w=j)],
\end{align*}
Where $\elbofinal_c$ denotes ELBO per component, $H[q(\hat{z}|w=c)$ denotes entropy per component and $q(\hat{\vz}|w=j)$ denotes entropy of variational distribution. 
% The natural gradient of the ELBO is the natural gradient of each summand scaled by $1/\pi_c$.

\section{Results}
The preliminary results in this section are obtained extending\footnote{\url{https://github.com/fmahdisoltani/multimodal_madam}} the Pytorch-SSO codebase of \cite{osawa2019practical}.\\More related prototyping results an be found in \url{https://github.com/fmahdisoltani/Variational_Inference}.
% \includemovie{1cm}{1cm}{images/h10_gmm2_mc1.gif}
% \includegraphics{images/h10_gmm2_mc1.jpg}

\begin{table}[h]
\begin{tabular}{cc}
1 component & 2 components \\
\includegraphics[height=2cm,width=5.5cm]{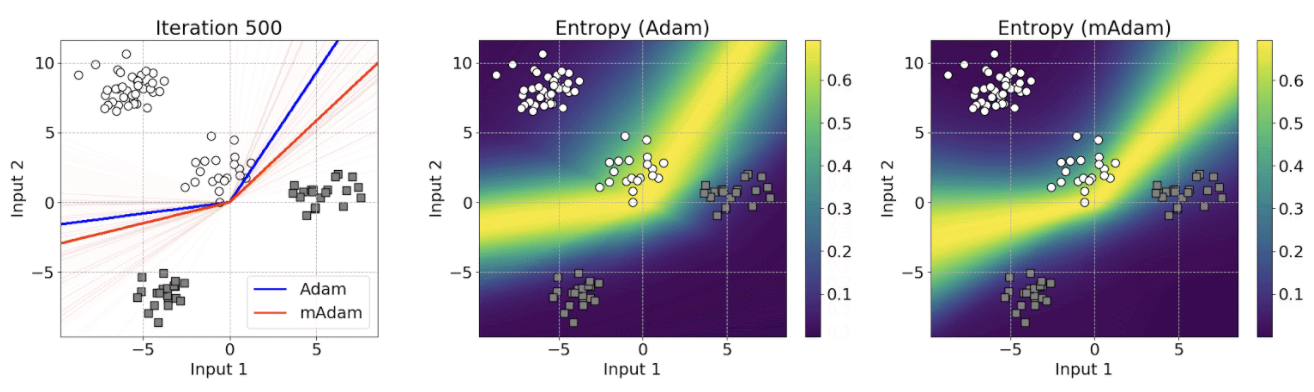} &
\includegraphics[height=2cm, width=5.5cm]{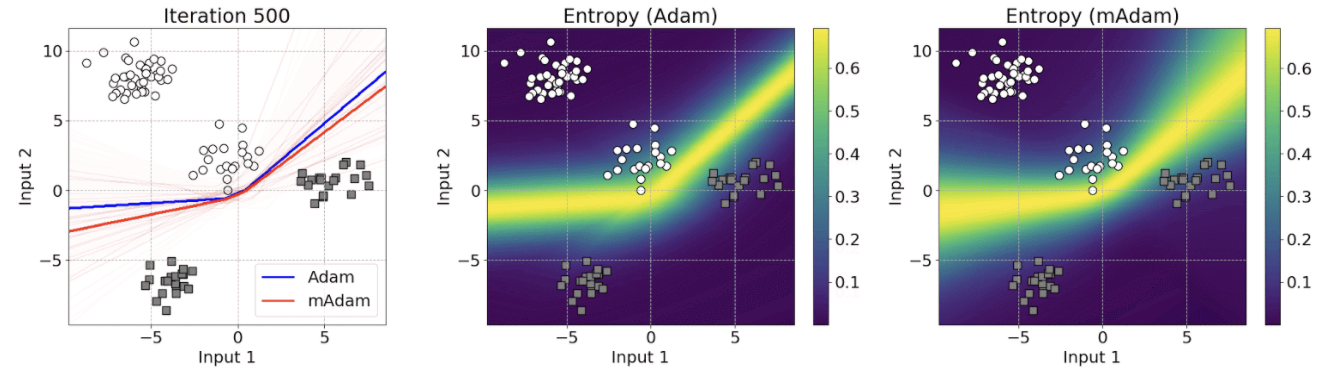} \\
\end{tabular}\\
\begin{tabular}{cc}
3 components & 4 components \\
\includegraphics[height=2cm,width=5.5cm]{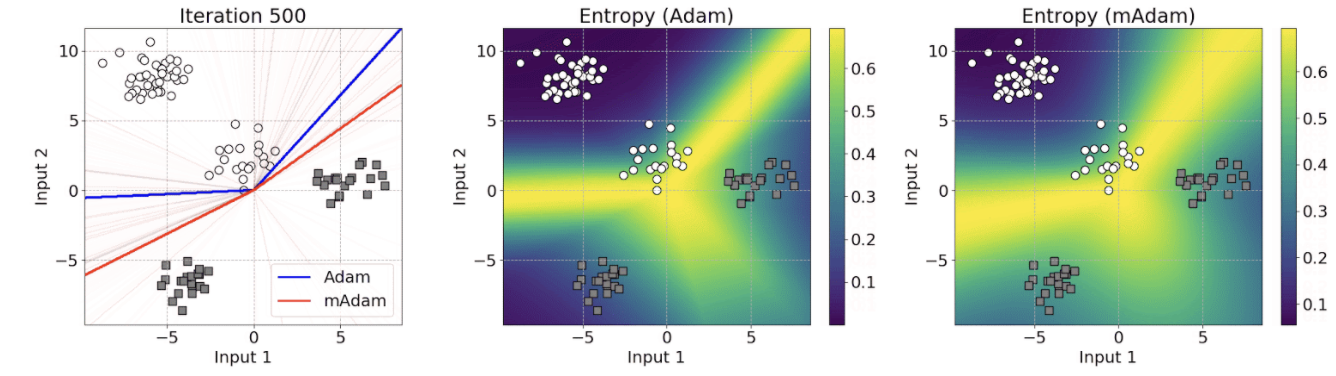} &
\includegraphics[height=2cm, width=5.5cm]{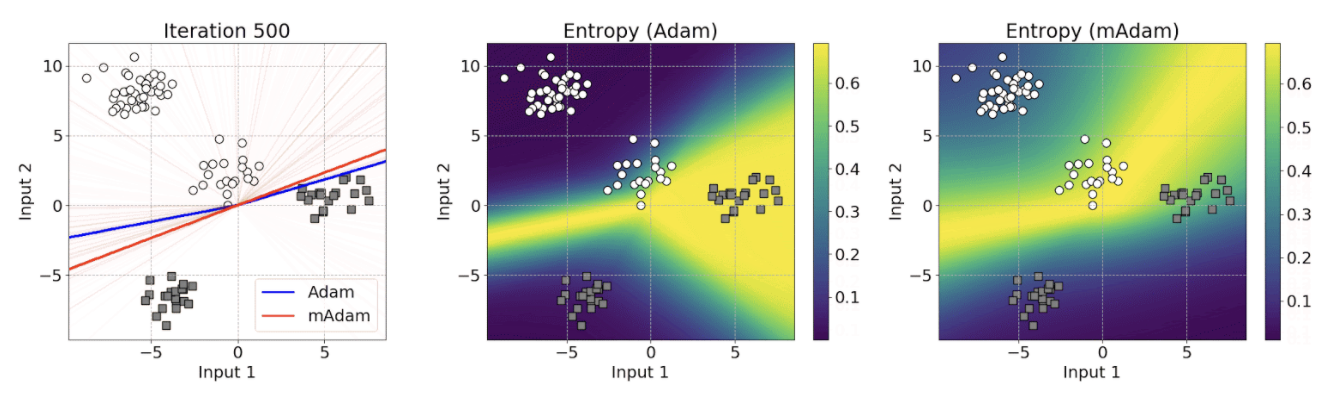} \\
\end{tabular}\\

\begin{tabular}{cc}
5 components \\
\includegraphics[height=2cm,width=5.5cm]{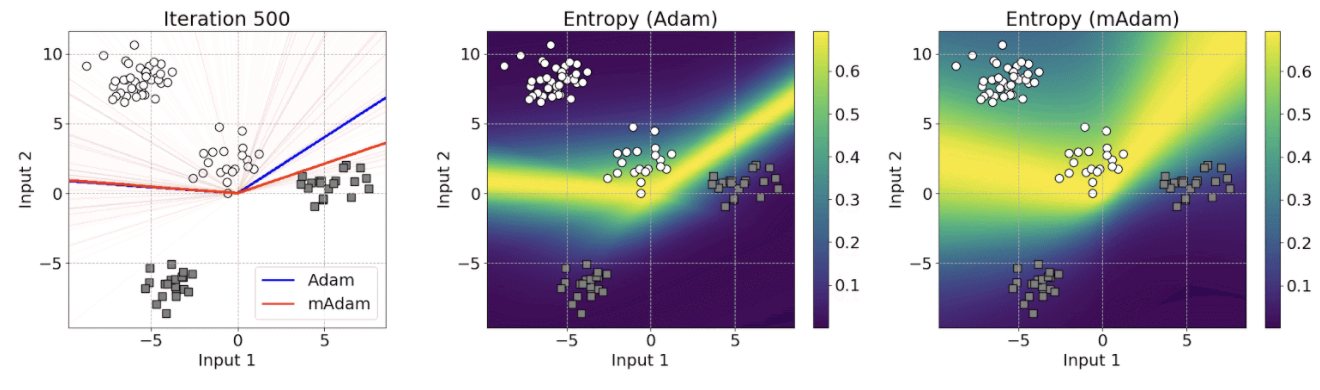}
\end{tabular}\\
\caption{The decision boundary obtained after 500 epochs, with 100 MC samples}
\end{table}
 
\begin{table}[h]
\begin{tabular}{c}
\includegraphics[height=4cm,width=11cm]{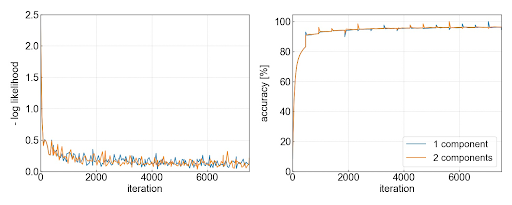}
\end{tabular}\\
\caption{The accuracy and log-likelihood for MNIST dataset vs iteration}
\end{table}

\bibliography{main}
% \clearpage

% \nocite{*} 
% \input{main.bbl} 

\end{document}